\documentclass[runningheads]{llncs}
\usepackage[T1]{fontenc}
\usepackage{graphicx}
\usepackage{amsmath}
\usepackage{amsfonts}
\usepackage{amssymb}
\usepackage{algorithmic}
\usepackage{textcomp}
\usepackage{xcolor}
\usepackage{algorithm}
\usepackage{array}
\usepackage{textcomp}
\usepackage{stfloats}
\usepackage{url}
\usepackage{verbatim}
\usepackage{makecell}
\usepackage{autobreak}
\usepackage{multirow}
\usepackage{multicol}
\usepackage[capitalize,noabbrev]{cleveref}
\usepackage{mathtools}
\usepackage{booktabs} 
\usepackage{enumitem}
\usepackage{setspace}
\usepackage{subfigure}

\usepackage{wrapfig}
\usepackage{caption}
\usepackage{subcaption}

\def\Mat#1{{\boldsymbol{#1}}}

\newcommand{\E}{\mathbb{E}}

\def\wrt{w.r.t}

\def\sB{{\mathcal{B}}}
\def\sI{{\mathcal{I}}}

\begin{document}
\title{Tackling the Local Bias in Federated Graph Learning}
%
%
\author{Binchi Zhang\inst{1,2}\thanks{Email: \texttt{epb6gw@virginia.edu}} \and
Minnan Luo\inst{1} \and
Shangbin Feng\inst{1,3} \and 
Ziqi Liu\inst{4} \and
Jun Zhou\inst{4} \and
Qinghua Zheng\inst{1}
}
\authorrunning{Binchi Zhang et al.}
%
\institute{Xi'an Jiaotong University \and
University of Virginia \and
University of Washington \and
Ant Group
}
%
\maketitle              
\begin{abstract}
Federated graph learning (FGL) has become an important research topic in response to the increasing scale and the distributed nature of graph-structured data in the real world.
In FGL, a global graph is distributed across different clients, where each client holds a subgraph. 
Existing FGL methods often fail to effectively utilize cross-client edges, losing structural information during the training; additionally, local graphs often exhibit significant distribution divergence. 
These two issues make local models in FGL less desirable than in centralized graph learning, namely the \textit{local bias} problem in this paper. 
To solve this problem, we propose a novel FGL framework to make the local models similar to the model trained in a centralized setting.
Specifically, we design a distributed learning scheme, \textbf{fully leveraging} cross-client edges to aggregate information from other clients.
In addition, we propose a \textbf{label-guided} sampling approach to alleviate the imbalanced local data and meanwhile, distinctly reduce the training overhead.
Extensive experiments demonstrate that local bias can compromise the model performance and slow down the convergence during training. 
Experimental results also verify that our framework successfully mitigates local bias, achieving better performance than other baselines with lower time and memory overhead.

\keywords{Federated graph learning \and Local bias \and Graph sampling.}
\end{abstract}
\section{Introduction}
Graph data has found increasing use in various real-world applications, such as social network analysis~\cite{fan2019graph,feng2022twibot}, protein structure prediction~\cite{fout2017protein,li2021structure}, and recommender systems~\cite{ying2018graph,wu2022graph}. 
However, many of these applications generate billions of nodes and edges, and the data is distributed across a large number of clients, which brings computational challenges when training graph learning models. 
Additionally, privacy concerns prevent local graph data from being collected to a central server during training. 
The above situations require federated graph learning (FGL), training a shared model while keeping local data decentralized~\cite{xie2021federated,zheng2021asfgnn,he2021fedgraphnn,wu2021fedgnn}. 
Existing FGL algorithms mainly address two types of problems: \emph{graph-level} problems and \emph{subgraph-level} problems~\cite{he2021fedgraphnn}. 
In graph-level problems, a large number of full graphs (e.g., molecular graphs) are stored in different clients. In subgraph-level problems, a global graph (e.g., a social network) is partitioned into different subgraphs (called \emph{local graphs}) stored in a distributed manner. 
This paper focuses on the latter setting.
In subgraph-level FGL problems, the main challenge centers around transmitting local graph data between different clients~\cite{fu2022federated}. 
Most graph learning models, such as graph neural networks (GNNs), learn powerful node representations by aggregating and transforming information from their neighbors. 
However, in subgraph-level FGL, many nodes in \emph{local graphs} may have neighbors stored in other clients, namely \emph{cross-client neighbors}. 
Due to privacy concerns, these nodes cannot directly aggregate information from their cross-client neighbors in the learning process. 
Existing works pursue different ways to tackle this problem.
In several existing works~\cite{he2021fedgraphnn,wu2021fedgnn}, if a node has cross-client neighbors, the client on which this node resides can request and collect embeddings of cross-client neighbors for this node from corresponding clients in a privacy-preserving manner. 
This way of aggregating neighbors is named \emph{neighbor collection} in this paper.
For other works~\cite{zhang2021subgraph,peng2022fedni}, nodes can learn the embeddings of their cross-client neighbors using a generative model trained on the local graphs, namely \emph{neighbor generation}.

Although these methods address the difficulty of transmitting local graph data, they do not achieve desirable performance as centralized graph learning models for the following reasons.
First, previous methods did not \textit{fully leverage cross-client neighbors}, so they cannot aggregate complete information as centralized graph learning methods on the global graph. 
In the neighbor collection, nodes can only aggregate information from other clients via its \textit{one-hop} cross-client neighbors; in the neighbor generation, the results of cross-client neighbor generation are not always reliable. 
In addition, the majority of existing FGL algorithms assume that local graphs in different clients are drawn from the same distribution~\cite{xie2021federated}. 
However, in practice, local graphs can differ significantly in size and exhibit substantial variation in distributions, leading to severe local model divergence and therefore degrading the performance of the global model.
These issues result in the local bias challenge for subgraph-level FGL problems.
To address this problem, FGL algorithms should be able to fully utilize cross-client neighbors during the training phase. Despite this, none of the existing FGL algorithms have successfully solved the local bias challenge. 

\begin{figure*}[t]
    \captionsetup{font=footnotesize}
    \centering
    \includegraphics[width=0.85\linewidth]{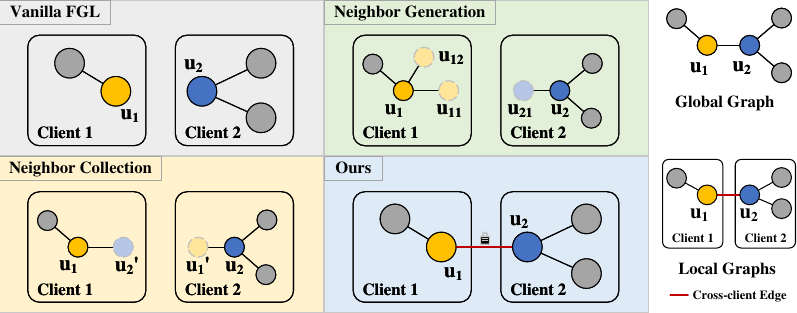}
    \vspace{-2mm}
    \caption{The comparison of our proposed framework with existing FGL frameworks on a toy example where the global graph is stored on two clients. Previous frameworks cut off the cross-client edge between Client 1 and Client 2. Our framework leverages the cross-client edges in the learning process while preserving data privacy.}
    \label{fig:difference}
    \vspace{-5mm}
\end{figure*}

In this paper, we propose an FGL framework that fully leverages the cross-client edges to tackle the local bias. 
Specifically, we reformulate the centralized graph learning process in a distributed learning paradigm.
In such a way of global training, our proposed FGL framework can achieve a training result similar to centralized graph learning, and consequently, the local bias issue is mitigated.
We show the difference between our method and existing FGL frameworks in \Cref{fig:difference}.
Vanilla FGL methods simply drop the cross-client edges, i.e. $(u_1,u_2)$, which corrupts the global graph structure. 
Neighbor generation algorithms exploit generated cross-client neighbors ($u_{11}$ and $u_{12}$) to aggregate cross-client information.
However, the generated neighbors can be different from the real cross-client neighbors ($u_2$).
Neighbor collection algorithms expand the local graphs by copying the cross-client neighbors of nodes in a privacy-preserving manner, e.g., $u_2^{\prime}$ with the same features as $u_2$ is added to the local graph in Client 1. 
However, neighbor collection can only expand the one-hop cross-client neighbors, while the insufficient aggregation of information from multi-hop cross-client neighbors can still lead to local bias.
Finally, our proposed framework allows local models to fully aggregate information from cross-client neighbors, exactly like centralized graph learning models.
Meanwhile, we propose a label-guided sampling approach to balance the size and class distribution of local graphs and reduce communication overhead.
Specifically, we first sample a local node subset in each client based on a label-guided probability. 
All these local subsets form a global subgraph which is then used for training.
Furthermore, we prove that our proposed label-guided sampling is unbiased and discuss the convergence of our framework. 
Our contributions are summarized as follows:
\begin{itemize}[leftmargin=*]
\item We propose a novel FGL framework, fully leveraging cross-client edges to aggregate global information and mitigate the local bias. 
\item We propose a label-guided subgraph sampling approach. This unbiased sampling technique alleviates local bias and improves the scalability of our method.
\item We conduct extensive experiments on three benchmark datasets. 
The results demonstrate that local bias hurts the model performance in heavy data decentralization and slows down the convergence, and verify that our method successfully mitigates the local bias.
\end{itemize}

\section{Proposed Method}\label{sec:proposed method}
\begin{figure*}[t]
    \captionsetup{font=footnotesize}
    \centering
    \includegraphics[width=0.85\linewidth]{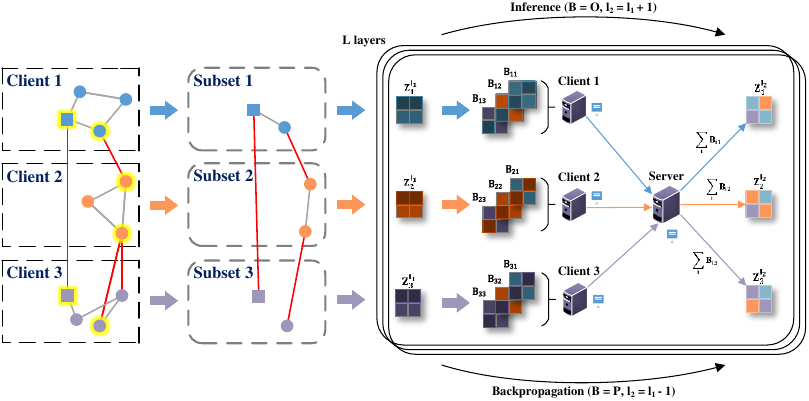}
    \vspace{-2mm}
    \caption{Illustration of our FGL framework where red edges denote the cross-client edges, $\Mat{Z}$ denotes the hidden matrix, $\Mat{O}$ and $\Mat{P}$ denote the local terms in the inference stage and the backpropagation stage, respectively. First, each client independently samples a node subset based on the label distribution, which forms a global subgraph across all clients. Then, each client performs our proposed distributed inference method with our communication scheme for $L$ layers and obtains the output. Finally, distributed backpropagation is conducted similarly to finish the training process.}
    \label{fig:model architecture}
    \vspace{-5mm}
\end{figure*}

Following commonly used notations, we use square brackets to represent the corresponding partitions in a matrix, e.g., $\Mat{X}[u,v]$ denotes the element in the $u^{th}$ row and $v^{th}$ column of matrix $\Mat{X}$. Furthermore, we use colons to represent the corresponding row or column in a matrix, e.g., $\Mat{X}[u,:]$ and $\Mat{X}[:,v]$ denote the $u^{th}$ row and the $v^{th}$ column of matrix $\Mat{X}$, respectively.
To facilitate better comprehension, we present some basic concepts utilized in this article. 
Let $\mathcal{G}=(\mathcal{V}, \mathcal{E})$ be a global undirected graph with $n$ nodes $v_i\in\mathcal{V}$ and edges $e_{ij}=(v_i, v_j)\in\mathcal{E}$ for $i,j=1,\dots,n$.
The adjacency matrix $\Mat{A}\in\mathbb{R}^{n\times n}$ indicates the connections between the edges. In a distributed setting, we assume that the global graph $\mathcal{G}$ is stored in $m$ remote clients $C_1,\dots,C_m$. 
The $i^{th}$ client $C_i$ has a subgraph $\mathcal{G}_i=\{\mathcal{V}_i,\mathcal{E}_i\}$, namely a \emph{local graph}, where $\mathcal{V}_i$ collects $n_i$ local nodes such that $\sum_{i=1}^mn_i=n$; $\mathcal{E}_i$ represents the edges between nodes in $\mathcal{V}_i$. 
In this way, each client $C_i$ is associated with a corresponding feature matrix $\Mat{X}_i\in\mathbb{R}^{n_i\times d}$ and a label matrix $\Mat{Y}_i\in\mathbb{R}^{n_i\times c}$, where $d$ and $c$ represent the dimension of the features and the number of classes of nodes, respectively.
According to the partition of $m$ clients, the adjacency matrix $\Mat{A}\in\mathbb{R}^{n\times n}$, feature matrix $\Mat{X}\in\mathbb{R}^{n\times d}$, and label matrix $\Mat{Y}\in\mathbb{R}^{n\times c}$ are partitioned as
\begin{equation}\label{eq:partition}
\small
\Mat{A}=
\begin{bmatrix}
\Mat{A}_{11} & \cdots & \Mat{A}_{1m} \\
\vdots & \ddots & \vdots \\
\Mat{A}_{m1} & \cdots & \Mat{A}_{mm} \\
\end{bmatrix}
,\ \Mat{X}=
\begin{bmatrix}
\Mat{X}_1 \\
\vdots \\
\Mat{X}_m \\
\end{bmatrix}
,\ \Mat{Y}=
\begin{bmatrix}
\Mat{Y}_1 \\
\vdots \\
\Mat{Y}_m \\
\end{bmatrix},
\end{equation}
where $\Mat{A}_{ij}$ is formed by $\mathcal{V}_i$'s corresponding rows and $\mathcal{V}_j$'s corresponding columns of adjacency matrix $\Mat{A}$. Specifically, we have $\Mat{A}_{ij}=\Mat{A}_{ji}^\top$. The client $C_i$ stores the local feature data $\Mat{X}_i$, and all edges (inner edges and cross-client edges) related to local nodes $\mathcal{V}_i$: $\{\Mat{A}_{i1},\cdots,\Mat{A}_{im}\}$.
We next introduce two parts contained in our method, a distributed model backbone and a label-guided sampling technique. 
We illustrate the overall framework in \Cref{fig:model architecture}.

\subsection{Model Backbone}
The goal of graph representation learning is to learn an encoder $f_{\Mat{W}}: \mathbb{R}^{n\times d}\times\mathbb{R}^{n\times n}\rightarrow\mathbb{R}^{n\times c}$, such that the high-level representations of the nodes in the $i^{th}$ client are obtained by $\Mat{H}_i=f_{\Mat{W}}\left(\Mat{X},\Mat{A}\right)$ for $i=1,\dots,m$, where the model parameter $\Mat{W}$ is shared by all clients. 
Here, we use GCN~\cite{kipf2016semi}, a mostly adopted graph learning model, as a specific example of $f_{\Mat{W}}$. 
It is worth noting that our discussion can be easily generalized to any GNN that is in matrix form, e.g., k-GNN~\cite{morris2019weisfeiler}, GIN~\cite{xu2018powerful}, and APPNP~\cite{gasteiger2018predict}.
An $L$-layer GCN model on the global graph $\mathcal{G}$ can be formulated as $\Mat{Z}^l=\sigma(\Mat{H}^l),\ \Mat{H}^l=\Mat{L}\Mat{Z}^{l-1}\Mat{W}^l$, for $l=1,2,\dots,L$, where $\sigma$ is an adaptable activation function, e.g., \emph{ReLU}; $\Mat{W}^l\in\mathbb{R}^{d^{l-1}\times d^l}$ collects the parameters in the $l^{th}$ layer; $d^l$ is the dimension of hidden representation of the $l^{th}$ layer $(l=0,\dots,L)$ such that $d^0=d$ and $d^L=c$. $\Tilde{\Mat{D}}=\Mat{D}+\Mat{I}$ refers to the diagonal degree matrix of the global graph $\mathcal{G}$ with self-loops.
$\Mat{L}=\sqrt{\Tilde{\Mat{D}}}\Tilde{\Mat{A}}\sqrt{\Tilde{\Mat{D}}}$ is the normalized Laplacian matrix.
$\Mat{Z}^l$ denotes the node embedding matrix after the $l^{th}$ layer. Specifically, we have $\Mat{Z}^0=\Mat{X}$.
We partition the matrices in centralized GNNs based on \Cref{eq:partition} and reformulate centralized GNNs in a distributed form as
\begin{equation}\label{eq:origin distributed GCN}
\small
\Mat{Z}^{l}_i=\sigma(\Mat{H}^l_i),\quad\Mat{H}^l_i=\sum_{j=1}^m\Mat{L}_{ij}\Mat{Z}^{l-1}_j\Mat{W}^l=\Mat{L}_{ii}\Mat{O}_i^{l-1}+\sqrt{\Tilde{\Mat{D}}_i}\sum_{j\neq i}\Tilde{\Mat{L}}_{ij}\Mat{O}_j^{l-1},
\end{equation}
for $l=1,2,\dots,L$, where $\Mat{Z}^l_i$ denotes the node embedding matrix on client $C_i$ after the $l^{th}$ layer; $\Mat{L}_{ij}$ is a block of the normalized Laplacian matrix $\Mat{L}=\sqrt{\Tilde{\Mat{D}}}\Tilde{\Mat{A}}\sqrt{\Tilde{\Mat{D}}}$ corresponding to the rows of nodes in $\mathcal{V}_i$ and columns of nodes in $\mathcal{V}_j$, where $\Tilde{\Mat{A}}=\Mat{A}+\Mat{I}$. 
We rewrite $\Mat{L}_{ij}$ as $\Mat{L}_{ij}=\sqrt{\Tilde{\Mat{D}}_i}\cdot\Tilde{\Mat{L}}_{ij}$ where $\Tilde{\Mat{L}}_{ij}=\Tilde{\Mat{A}}_{ij}\sqrt{\Tilde{\Mat{D}}_j}$. 
Consequently, the local term $\Tilde{\Mat{L}}_{ij}\Mat{O}_j^{l-1}$ is computed within the client $C_j$ and sent to the server. 
Finally, the collected local terms are summed up in the server, and the result $\sum_{j\neq i}\Tilde{\Mat{L}}_{ij}\Mat{O}_j^{l-1}$ is sent back to the client $C_i$. 

Given a global graph $\mathcal{G}$ with $n$ nodes scattered in $m$ clients, our distributed training objective is to optimize a weighted sum of loss over $m$ clients, i.e.,
\begin{equation}\label{eq:obj}
\small
\min_{\Mat{W}}\ \mathcal{L}\left(\Mat{W}\right)=\sum_{i=1}^{m}p_iLoss_i(\Mat{Z}_i^L,\Mat{W}),
\end{equation}
where $\Mat{W}=\{\Mat{W}^1,\dots,\Mat{W}^L\}$ collects the parameters in all layers. $p_i\geq 0$ refers to the weight coefficient of the $i^{th}$ client $C_i$. $Loss_i$ is a task-specific local loss function, e.g., cross-entropy loss.
To solve the optimization problem (\ref{eq:obj}), we design a distributed backpropagation process.
It is obvious that the gradient of loss $\mathcal{L}$ with respect to the representation of the $L^{th}$ layer $\Mat{Z}^L_i$ can be calculated locally in $C_i$.
Recursively, the gradients of loss function $\mathcal{L}$ with respect to $\Mat{Z}^l_i$ and $\Mat{W}^l$ are derived as
\begin{equation}\label{eq:Zl-1}
\small
\nabla_{\Mat{Z}^{l-1}_i}\mathcal{L}=\Mat{L}_{ii}\Mat{P}_i^l+\sqrt{\Tilde{\Mat{D}}_i}\sum_{j\neq i}\Tilde{\Mat{L}}_{ij}\Mat{P}_j^l,\quad \nabla_{\Mat{W}^l}\mathcal{L}=\sum_{j=1}^m\left(\Mat{Z}^{l-1}_j\right)^\top\nabla_{\Mat{Z}^{l-1}_j}\mathcal{L},
\end{equation}
for $l=1,\dots,L-1$, where $\Mat{P}_i^l=(\nabla_{\Mat{Z}^l_i}\mathcal{L}\circ\sigma^\prime(\Mat{H}_i^l))({\Mat{W}^l})^\top$, $\circ$ is the Hadamard product. 
$\nabla_{\Mat{Z}^{l-1}_i}\mathcal{L}$ can be computed in the same way as \Cref{eq:origin distributed GCN} regarding their identical forms and each summation term of $\nabla_{\Mat{W}^l}\mathcal{L}$ can be obtained locally.

\subsection{Label-Guided Sampling}
Given the local distribution divergence and training overhead of the distributed algorithm, we propose a label-guided distributed graph sampling approach to address these issues.
Assuming the conditional distributions $P(\Mat{X}_i|\Mat{Y}_i)$ are similar for different $i$, we independently sample $s_i$ nodes in each client $C_i$ according to a label-guided probability distribution $\Mat{Q}_i$, and collect them into a set $\mathcal{V}_{S_i}\subseteq\mathcal{V}_i$. 
Let $\mathcal{V}_S=\bigcup_{i=1}^m\mathcal{V}_{S_i}$ be the union of nodes sampled on all clients, and $p\left(v\right)=P\left(v\in\mathcal{V}_S\right)=1-\left(1-\Mat{Q}_i\left(v\right)\right)^{s_i}$ be the probability of $v\in\mathcal{V}_S$.
When local graphs are non-i.i.d., the label-guided probability $\Mat{Q}_i$ balances the number and the distribution of the sampled set $\mathcal{V}_{S_i}$ in client $C_i$. $\Psi:V\rightarrow\{1,\dots,c\}$ is the label mapping. We take $n_i^k$ as the number of nodes with the label $k\in\{1,\dots,c\}$ in $\mathcal{V}_i$ and $n^k=\sum_{i=1}^mn_i^k$. 
For larger $n_j$, the sampling size $s_j$ should be larger to reduce the variance. Therefore, when the size of the global subgarph $\mathcal{V}_S$ is set as $s=\sum_{i=1}^ms_i=|\mathcal{V}_S|$, we let $s_j=\frac{n_j\cdot s}{n}$.
Then, to ensure the unbiasedness of our subgraph sampling method, we let $Q_i\left(u\right)=\frac{n^{\Psi\left(u\right)}}{n_i^{\Psi\left(u\right)}\sum_{k\in\Psi\left(\mathcal{V}_i\right)}n^k}$ and exploit probability $p\left(v\right)=1-\left(1-\Mat{Q}_i\left(v\right)\right)^{s_i}$ to normalize the hidden node embedding $\Mat{H}_i^l\left[v,:\right]$ for $v\in\mathcal{V}_{S_i}$ as
\begin{equation}
\label{eq:subgraph estimator}
\small
\Mat{H}_{S_i}^l\left[v,:\right]=\sum_{u\in\mathcal{V}_{S_i}}\frac{\Mat{L}_{ii}\left[v,u\right]}{p\left(u\right)}\Mat{O}_{S_i}^{l-1}\left[u,:\right]+\sqrt{\Tilde{\Mat{D}}_i\left[v,v\right]}\sum_{j\neq i}\sum_{u\in\mathcal{V}_{S_j}}\frac{\Tilde{\Mat{L}}_{ij}\left[v,u\right]}{p\left(u\right)}\Mat{O}_{S_j}^{l-1}\left[u,:\right],
\end{equation}
where $\Mat{O}_{S_i}^{l}=\sigma(\Mat{H}_{S_i}^l)\Mat{W}^{l+1}$.
Consequently, we obtain $\Hat{\Mat{H}}_{S_i}^l\in\mathbb{R}^{s_i\times d^l}$ as an estimator of the hidden representation matrix $\Mat{H}_i^l$. 
In addition, $\Hat{\Mat{H}}_{S_i}^l$ is proven to be an unbiased estimator.
The proof can be found in Appendix B.
\begin{proposition}\label{thr:proposition1}
Matrix $\Mat{H}_{S_i}^l$ defined in \Cref{eq:subgraph estimator} is an unbiased estimator of hidden representation matrix $\Mat{H}_i^l$, i.e., $\E[\Mat{H}_{S_i}^l]=\Mat{H}_i^l$.
\end{proposition}
Correspondingly, the optimization problem (\ref{eq:obj}) is changed to $\min_{\Mat{W}}\ \mathcal{L}_S(\Mat{W})=\sum_{i=1}^mp_iLoss_{i}(\Mat{Z}_{S_i}^L,\Mat{W})$.
It is worth noting that sampling can lose some cross-client edges during training. 
However, our framework is equivalent to graph sampling-based centralized GNNs~\cite{chiang2019cluster,zeng2019graphsaint}, greatly alleviating the local bias issue. 

\begin{table*}[t]
\captionsetup{font=footnotesize}
\footnotesize
\centering
\tabcolsep = 3 pt
\caption{Time and communication cost in the forward step and memory cost in each iteration of different models. For CAGNET \cite{tripathy2020reducing}, we choose the method with the least communication overhead, that is, the 1.5d version. $c<m$ is a partitioning factor; $t$ is the number of sampled neighbors for each node and $t_b$ is the average number of boundary nodes in sampled neighbors; $n_S$ is the size of the sampled subgraph/layer; $n_{\sB}$ is the size of boundary node set; $n_{\sB S}$ is the number of boundary nodes in the sampled subgraph; $n_{\sI S}$ is the number of inner nodes in the sampled subgraph.}\label{tab:comm_cost}
\begin{tabular}{lcccc}
    \toprule[1pt]
    & CAGNET & FedAvg + GCN & BNS-GCN & Ours \\
    \midrule[0.5pt]
    Time & $\frac{2}{m}nLd^2+\frac{n}{m}d$ & $\frac{2}{m}nLd^2+\frac{n}{m}d$ & $2n_SLd^2+n_Sd$ & $2n_SLd^2+n_Sd$ \\
    Comm & $\frac{n}{c}Ld+\frac{c}{m}nLd$ & $mLd^2$ & $n_{\sB S}Ld$ & $mn_SLd$ \\
    Memory & $\frac{c}{m}nLd+cLd^2$ & $\frac{n}{m}Ld+mLd^2$ & $n_SLd+mLd^2+2n_{\sB S}Ld$ & $n_SLd+mLd^2$ \\
    \bottomrule[1pt]
\end{tabular}
\vspace{-5mm}
\end{table*}


\section{Analysis}
\subsection{Overhead}\label{sec:overhead}
We compare our proposed framework with the existing distributed GCN algorithms in \Cref{tab:comm_cost}. As \Cref{tab:comm_cost} shows, we typically have ours $=$ BNS-GCN $\leq$ FedAvg+GCN $=$ CAGNET for computation time, ours $\leq$ BNS-GCN $\leq$ FedAvg+GCN $\leq$ CAGNET for memory overhead, and FedAvg + GCN $\leq$ BNS-GCN $\leq$ ours $\leq$ CAGNET for communication cost. 
From the results in \Cref{tab:comm_cost}, we achieve the lowest overhead of memory and computation time (BNS-GCN can be seen as a specific case of ours where $p_v=1$ when $v\in\sB$ and $p_v=\frac{1}{n_S}$ otherwise). 
Furthermore, the sampling technique greatly reduces the time and memory overhead and makes the training more scalable in a distributed setting.
We provide a detailed complexity analysis in Appendix C.

\subsection{Convergence}
The convergence of our framework is guaranteed as follows.
\begin{theorem}\label{thr:proposition3}
Let $\E_T\|\nabla\mathcal{L}(\Mat{W}_T)\|_2=\frac{1}{T}\sum_{t=0}^{T-1}\|\nabla\mathcal{L}(\Mat{W}_t)\|_2$ where $T$ is the number of iterations. Assume $\mathcal{L}$ is $\rho$-smooth. Let $\eta\leq\frac{1}{\rho}$, we have $\E_T\Vert\nabla\mathcal{L}\left(\Mat{W}_T\right)\Vert_2^2\leq\frac{8\left(\mathcal{L}\left(\Mat{W}_0\right)-\mathcal{L}\left(\Mat{W}_T\right)\right)}{T\eta\left(2-\rho\eta\right)}$ with probability $1$.
\end{theorem}
The proof of~\Cref{thr:proposition3} is shown in Appendix B. 
\Cref{thr:proposition3} indicates that the optimization problem~(\ref{eq:obj}) can be solved at a local optimum (with a bounded gradient) by our proposed framework with label-guided sampling.

\subsection{Privacy}
The main difference between our framework and previous FGL frameworks is we leverage the cross-client information during the local inference.
We argue that our framework does not breach the requirements in FL.
For client $C_i$, we only upload the products $\Tilde{\Mat{L}}_{S_{ij}}\Mat{O}_{S_j}^l$ and $\Tilde{\Mat{L}}_{S_{ij}}\Mat{P}_{S_j}^l$ to the central server instead of the raw data $\Mat{X}_i$ and $\Mat{A}_{ij}$, where $\Tilde{\Mat{L}}_{S_{ij}}$ is the block corresponding to $\mathcal{V}_{S_i}$ and $\mathcal{V}_{S_j}$.
Because the terms $\Mat{L}_{S_{ii}}\Mat{O}_{S_i}^{l-1}$ and $\Mat{L}_{S_{ii}}\Mat{P}_{S_i}^l$ are stored locally on the client $C_i$, the central server is not able to recover the hidden matrices or their gradients with the uploaded data according to \Cref{eq:origin distributed GCN} and \Cref{eq:Zl-1}.
In addition, the local calculation of the client $C_i$ only requires local structural information $\Mat{A}_{ij}$ and $\Mat{D}_i$.
It is worth noting that $\Mat{A}_{ij}$ is naturally visible to the client $C_i$ and $C_j$.

\section{Experiments}

\subsection{Experimental Setup}
In this section, we conduct empirical studies on the problem of local bias and evaluate our proposed method.
Specifically, our objective is to answer the following research questions through our experiments: 
\textbf{RQ1:} Can incomplete utilization of cross-client neighbors and imbalanced local distribution lead to local bias?
\textbf{RQ2:} Can our proposed training framework and label-guided sampling methods mitigate the local bias problem? 
\textbf{RQ3:} How does our proposed sampling method benefit federated graph learning on efficiency?
We conduct experiments on three prevalent graph datasets, PubMed, Reddit, and Ogbn-products.
More details on experimental settings and supplementary results are in Appendix D and E.



\subsection{Experimental Results}
\begin{wraptable}[15]{r}{0.6\textwidth}
\captionsetup{font=footnotesize}
\footnotesize
\vspace{-8mm}
\caption{Micro F1-score of different methods on three datasets. "OOM" indicates that the corresponding case results in an out-of-memory issue.}\label{tab:f1_micro}
\vspace{-3mm}
\aboverulesep = 0pt
\belowrulesep = 0pt
\begin{tabular}{lccc}
    \toprule[1pt]
    Methods & PubMed & Reddit & Ogbn-products \\
    \midrule[0.5pt]
    GCN~\cite{kipf2016semi} & 87.80\% & 93.30\% & 75.64\% \\
    1-GNN~\cite{morris2019weisfeiler} & 90.50\% & OOM & OOM \\
    GraphSAGE~\cite{hamilton2017inductive} & 90.10\% & 95.40\% & 78.50\% \\
    GraphSAINT~\cite{zeng2019graphsaint} & 89.20\% & \textbf{96.60\%} & 79.08\% \\
    \midrule[0.5pt]
    FedAvg+GCN & 87.70\% & 91.86\% & 71.51\% \\
    CAGNET~\cite{tripathy2020reducing}& 88.20\% & 94.13\% & 75.36\% \\
    BNS-GCN~\cite{wan2022bns} & 90.20\% & 96.30\% & 78.66\% \\
    FedSAGE+~\cite{zhang2021subgraph} & 89.05\% & OOM & OOM \\
    LLCG~\cite{ramezani2021learn} & 89.20\% & 96.47\% & 74.00\% \\
    \midrule[0.5pt]
    Ours+GCN & 89.90\% & 95.54\% & \textbf{79.15\%} \\
    Ours+1-GNN & \textbf{90.70\%} & 96.12\% & 76.28\% \\
    \bottomrule[1pt]
\end{tabular}
\end{wraptable}
We compare the basic performance of our proposed framework and the baselines in three datasets in \Cref{tab:f1_micro}.
We use GCN \cite{kipf2016semi} and 1-GNN \cite{morris2019weisfeiler} as the backbone of our framework.
The results in \Cref{tab:f1_micro} demonstrate that
(1) Our framework achieves the best performance on PubMed and Ogbn-products. 
(2) Our framework has competitive performance compared to centralized baselines, indicating our success in mitigating the local bias.
(3) Among all distributed baselines, FedAvg is the only one without using cross-client information, resulting in the worst performance on all datasets. This result verifies the importance of the cross-client edges in FGL.
(4) The results verify the effectiveness of sampling (GraphSAGE, GraphSAINT, BNS-GCN, and ours) on both the scalability and the model performance.



\begin{wrapfigure}[11]{r}{0.7\textwidth} 
\captionsetup[subfigure]{font=footnotesize}
\captionsetup{font=footnotesize}
\centering
\vspace{-9mm}
\subfigure[Comparison of f1-score under different clients numbers]{
\includegraphics[width=0.47\linewidth]{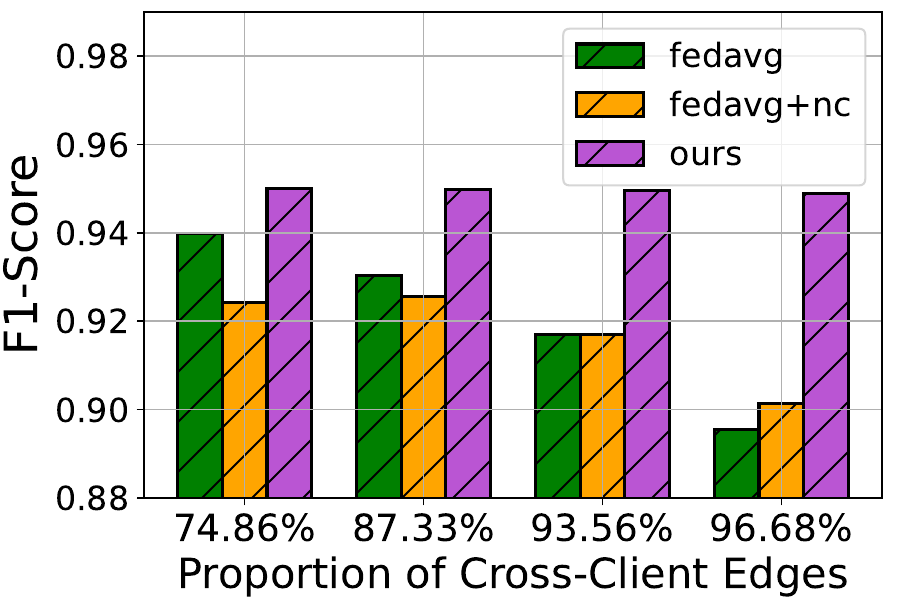}}
\subfigure[Comparison of local bias under different clients numbers]{
\includegraphics[width=0.47\linewidth]{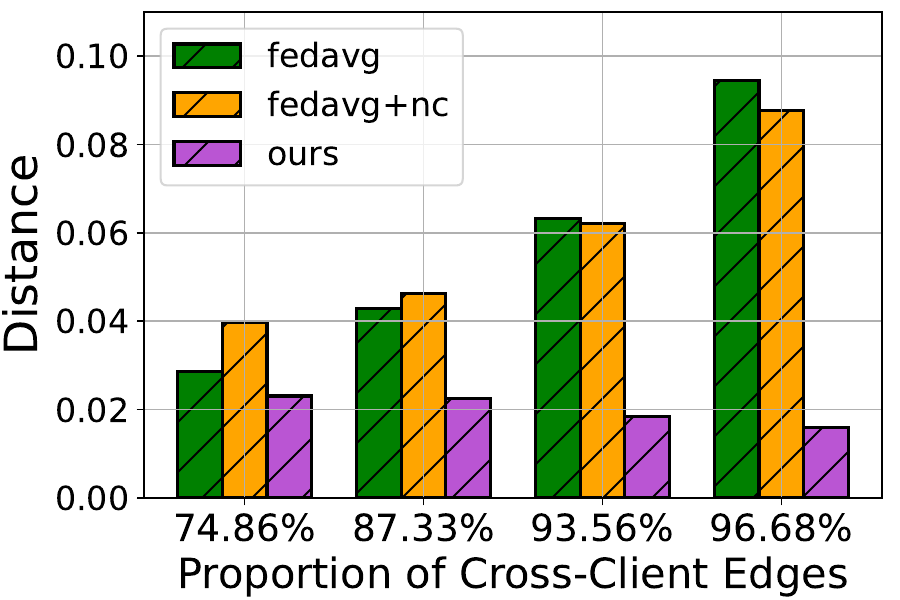}}
\vspace{-4mm}
\caption{Impact of cross-client edges on local bias.}
\label{fig:cross-client edges}
\end{wrapfigure}

To answer \textbf{RQ1} and \textbf{RQ2}, we empirically verify the effect of the complete utilization of cross-client edges on the local bias.
In particular, we compare our method (complete) with two baselines, FedAvg (none) and FedAvg with neighbor collection (incomplete).
For FedAvg with the neighbor collection, we randomly select 20\% cross-client neighbors for each client and add them to the local graph on the client.
To measure the extent of local bias, we use the average \textbf{Euclidean distance} of output logits of the FGL model and the centralized graph learning model according to the definition of local bias, where a smaller distance indicates a smaller local bias of the FGL model. 
Moreover, to highlight the impact of cross-client edges on the local bias, we change the proportion of cross-client edges by modifying the number of clients.
As the number of clients increases, the proportion of cross-clients grows larger, correspondingly.
The experimental results are shown in \Cref{fig:cross-client edges}.
We change the number of clients across 4, 8, 16, and 32, and the proportion of cross-client edges changes across 74.86\%, 87.33\%, 93.56\%, and 96.68\%, respectively.
We can observe that our method has a desirable performance under different proportions of cross-client edges.
In contrast, the performance of other baselines gets worse as the proportion of cross-client edges increases because of the effect of incompletely utilizing cross-client edges.
Additionally, we find that FedAvg and FedAvg+nc have an increasing local bias when the proportion of cross-client edges increases, indicating a positive answer to \textbf{RQ1}. 
It is worth noting that the local bias of our method decreases as the proportion of cross-client edges increases.
We explain this phenomenon as our model is trained on a global subgraph constructed by local nodes sampled from all clients; as the proportion of cross-client edges increases, the proportion of cross-client edges in the global subgraph also increases.
Consequently, our method utilizes more cross-client edges as their proportion increases.

In addition, we also investigate the effect of imbalanced local distributions on local bias.
We compare our method with the same baselines under two settings.
\begin{wrapfigure}[14]{r}{0.6\textwidth} 
\captionsetup[subfigure]{font=footnotesize}
\captionsetup{font=footnotesize}
\centering
\vspace{-8mm}
\subfigure[F1-score and local bias on normal and non-i.i.d. settings.]{
\includegraphics[width=0.47\linewidth]{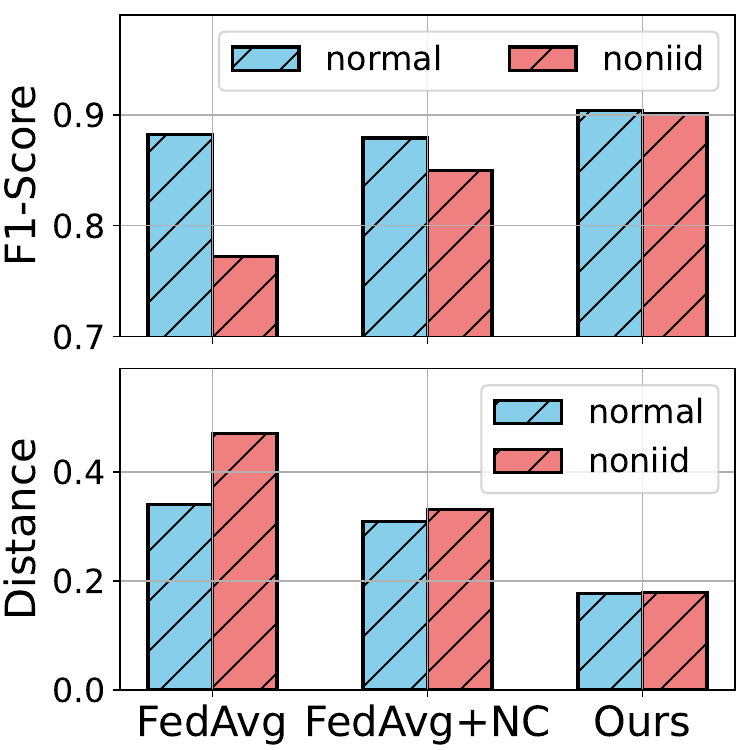}\label{fig:noniid}}
\subfigure[Convergence curves of baselines on normal and non-i.i.d. settings.]{
\includegraphics[width=0.47\linewidth]{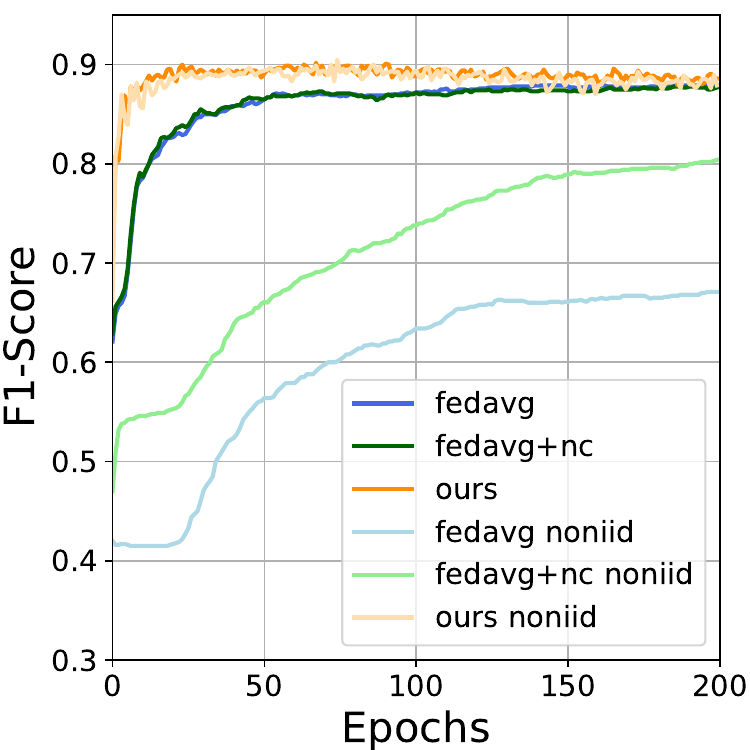}\label{fig:convergence}}
\vspace{-5mm}
\caption{The impact of imbalanced local distribution on the local bias problem.}
\label{fig:local noniid}
\end{wrapfigure}
In the normal setting, we randomly partition the local nodes based on a uniform distribution to ensure that the label distributions of different local graphs are similar. 
In the non-i.i.d. setting, we randomly choose a few classes from the label domain and make the sampling probability of the nodes in these classes \textit{ten times} more than the other classes in each client.
The experimental results are shown in \Cref{fig:local noniid}.
In \Cref{fig:noniid}, we compare the test f1 score of three baselines in normal and non-i.i.d. settings and observe that
(1) Without leveraging cross-client edges, the FedAvg method has the largest performance drop (over 10\%) under imbalanced local distributions.
(2) The neighbor collection technique can improve the performance of the FGL model under imbalanced local distributions.
However, there is still a distinct gap between the two settings.
(3) In comparison, our method only has a mere difference in the test f1 score between normal and non-i.i.d. settings.
For the local bias metric, we can observe that
(1) Our method has a lower local bias in both normal and non-i.i.d. settings. 
(2) Imbalanced local distribution will result in a larger local bias for the baselines. In contrast, our method is not affected by local distribution divergence due to the label-guided sampling technique.
(3) Complete utilization of the cross-client edges can mitigate the impact of local distribution divergence.
In \Cref{fig:convergence}, we compare the convergence curves of three baselines in normal and non-i.i.d. settings.
We find that local distribution divergence can distinctly reduce the convergence rate of two baselines.
In contrast, our method converges much faster and is unaffected by local distribution divergence.


\subsection{Training Overhead Study}
\begin{table*}[t]
\captionsetup{font=footnotesize}
\footnotesize
\centering
\tabcolsep = 3 pt
\caption{Communication overhead, local computation time, and memory overhead in an epoch. We adopt 5,000, 50,000, and 500,000 as the sampling size for PubMed, Reddit, and Ogbn-products, respectively.}
  \label{tab:time cost}
\aboverulesep = 0pt
\belowrulesep = 0pt
\begin{tabular}{l|ccc|ccc}
    \toprule[1pt]
    \multicolumn{1}{l|}{} & \multicolumn{3}{c|}{Training cost / s} & \multicolumn{3}{c}{Memory cost / MB} \\ 
    Method & PubMed & Reddit & Products & PubMed & Reddit & Products \\
    \midrule[0.5pt]
    CAGNET (4 Clients) & 0.3063 & 21.1358 & 38.3888 & 20.20 & 698.09 & 855.34 \\
    FedAvg+GCN (4 Clients) & 0.0937 & 12.0175 & 16.9388 & 10.65 & 285.00 & 402.27 \\
    Ours+GCN (4 Clients) & 0.0526 & 3.3379 & 5.0169 & 3.50 & 60.85 & 100.15 \\
    \midrule[0.5pt]
    CAGNET (8 Clients) & 0.4078 & 15.2789 & 35.9513 & 10.10 & 349.34 & 427.76 \\
    FedAvg+GCN (8 Clients) & 0.0994 & 4.5173 & 7.5080 & 5.46 & 104.38 & 163.44 \\
    Ours+GCN (8 Clients) & 0.1701 & 3.3461 & 4.3342 & 1.75 & 30.43 & 50.08 \\
    \bottomrule[1pt]
  \end{tabular}
\vspace{-5mm}
\end{table*}
To answer \textbf{RQ3}, we compare our time and memory overhead with other distributed methods in \Cref{tab:time cost}.
First, we compare our proposed method with a classic full-batch distributed GCN training method, CAGNET \cite{tripathy2020reducing} and the vanilla FedAvg algorithm. For \Cref{tab:time cost}, we use the \emph{time} function in Python to count the time cost in each epoch. The results demonstrate that our method involves consistently lower communication overhead in both 4- and 8-client settings. Especially in large datasets, e.g. Ogbn-products, our method reduces the training time by more than 80\% compared with CAGNET, for 50\% compared with the FedAvg algorithm. Although the communication increases a little with regard to FedAvg, the time for local computation exhibits a significant decrease, which yields lower training time costs even with heavy data decentralization.
Besides, we also examine whether our method reduces memory overhead in the training process. Memory cost consists primarily of three parts, node features in PyTorch computational graphs, model parameters, and sparse adjacency matrices, which are distributed in different clients. We record memory costs in one epoch for each client in \Cref{tab:time cost}. Our method without introducing redundant memory successfully reduces the memory cost compared with other distributed methods because of our sampling technique. Specifically, the memory cost of our method is 2/3 less than that of FedAvg.
In conclusion, our proposed label-guided sampling technique distinctly reduces both time and memory overhead.

\section{Related Works}

\paragraph{Distributed Training of GNNs.}
As the increasing scale of graph data poses challenges to training GNNs on a single machine, distributed training of GNNs has emerged as a promising solution.
Scardapane et al.~\cite{scardapane2020distributed} presented a fully distributed training framework with an optimization criterion for designing network topology in which the global graph is stored in distributed clients. 
Tripathy et al.~\cite{tripathy2020reducing} presented a family of parallel GCN training frameworks (CAGNET). They reduced communication by partitioning adjacency matrices (sparse) and feature matrices (dense) in different ways and assigning computations to different clients. 
Wan et al.~\cite{wan2022bns} developed graph sampling methods on boundary nodes that are responsible for communication and increase throughput by up to 500\%. 
In contrast, their solution stores node features in the boundary set multiple times, increasing the memory overhead.
Ramezani et al.~\cite{ramezani2021learn} introduced a local-learn-global-correct framework for distributed GNN training where the local models are refined by global server corrections before model averaging.

\paragraph{Subgraph-level Federated Graph Learning.}
To address data decentralization in graph-based applications, researchers incorporated federated learning into graph learning systems, resulting in the development of FGL \cite{fu2022federated,peng2022fedni,wu2021fedgnn}.
Zhang et al.~\cite{zhang2021subgraph} incorporated a missing neighbor generator to expand local graphs with generated cross-client neighbors without exchanges of raw data across clients.
Wang et al.~\cite{wang2020graphfl} introduced an FGL framework based on MAML~\cite{finn2017model}, which learned a global model among clients to address the non-i.i.d. local data and adapted to new label domains.
Zheng et al.~\cite{zheng2021asfgnn} presented an FGL model that exploits JS divergence to improve the local model and Bayesian optimization to tune the hyperparameters.  
Wu et al.~\cite{wu2021fedgnn} proposed a novel FGL model for recommendation. They develop a local user-item graph expansion to capture higher-order user-item interaction and a pseudo-interacted item sampling approach to protect user privacy.

\section{Conclusion}

In this paper, we found that existing subgraph-level FGL methods failed to fully utilize the cross-client edges while tackling the local distribution divergence, leading to the local bias problem.
To address this problem, we presented a novel FGL framework with global training and balanced subgraph sampling. 
We also provided theoretical analysis of our framework. 
Extensive experiments verified the harm of local bias and the efficacy of our framework in reducing local bias.

%
%
%
\bibliographystyle{splncs04}
\bibliography{pakdd2024}

\begin{thebibliography}{10}
\providecommand{\url}[1]{\texttt{#1}}
\providecommand{\urlprefix}{URL }
\providecommand{\doi}[1]{https://doi.org/#1}

\bibitem{chiang2019cluster}
Chiang, W.L., Liu, X., Si, S., Li, Y., Bengio, S., Hsieh, C.J.: Cluster-gcn: An efficient algorithm for training deep and large graph convolutional networks. KDD  (2019)

\bibitem{fan2019graph}
Fan, W., Ma, Y., Li, Q., He, Y., Zhao, E., Tang, J., Yin, D.: Graph neural networks for social recommendation. In: TheWebConf (2019)

\bibitem{feng2022twibot}
Feng, S., Tan, Z., Wan, H., Wang, N., Chen, Z., Zhang, B., Zheng, Q., Zhang, W., Lei, Z., Yang, S., et~al.: Twibot-22: Towards graph-based twitter bot detection. NeurIPS  (2022)

\bibitem{finn2017model}
Finn, C., Abbeel, P., Levine, S.: Model-agnostic meta-learning for fast adaptation of deep networks. In: ICML (2017)

\bibitem{fout2017protein}
Fout, A., Byrd, J., Shariat, B., Ben-Hur, A.: Protein interface prediction using graph convolutional networks. NeurIPS  (2017)

\bibitem{fu2022federated}
Fu, X., Zhang, B., Dong, Y., Chen, C., Li, J.: Federated graph machine learning: A survey of concepts, techniques, and applications. SIGKDD Explorations  (2022)

\bibitem{gasteiger2018predict}
Gasteiger, J., Bojchevski, A., G{\"u}nnemann, S.: Predict then propagate: Graph neural networks meet personalized pagerank. ICLR  (2019)

\bibitem{hamilton2017inductive}
Hamilton, W.L., Ying, R., Leskovec, J.: Inductive representation learning on large graphs. NeurIPS  (2017)

\bibitem{hamilton2017representation}
Hamilton, W.L., Ying, R., Leskovec, J.: Representation learning on graphs: Methods and applications. IEEE Data Eng. Bull.  (2017)

\bibitem{he2021fedgraphnn}
He, C., Balasubramanian, K., Ceyani, E., Rong, Y., Zhao, P., Huang, J., Annavaram, M., Avestimehr, S.: Fedgraphnn: A federated learning system and benchmark for graph neural networks. arXiv:2104.07145  (2021)

\bibitem{hu2020open}
Hu, W., Fey, M., Zitnik, M., Dong, Y., Ren, H., Liu, B., Catasta, M., Leskovec, J.: Open graph benchmark: Datasets for machine learning on graphs. NeurIPS  (2020)

\bibitem{doi:10.1137/S1064827595287997}
Karypis, G., Kumar, V.: A fast and high quality multilevel scheme for partitioning irregular graphs. SISC  (1998)

\bibitem{kingma2014adam}
Kingma, D.P., Ba, J.: Adam: A method for stochastic optimization. ICLR  (2015)

\bibitem{kipf2016semi}
Kipf, T.N., Welling, M.: Semi-supervised classification with graph convolutional networks. ICLR  (2017)

\bibitem{li2021structure}
Li, S., Zhou, J., Xu, T., Huang, L., Wang, F., Xiong, H., Huang, W., Dou, D., Xiong, H.: Structure-aware interactive graph neural networks for the prediction of protein-ligand binding affinity. In: KDD (2021)

\bibitem{mcmahan2017communication}
McMahan, B., Moore, E., Ramage, D., Hampson, S., y~Arcas, B.A.: Communication-efficient learning of deep networks from decentralized data. AISTATS  (2017)

\bibitem{morris2019weisfeiler}
Morris, C., Ritzert, M., Fey, M., Hamilton, W.L., Lenssen, J.E., Rattan, G., Grohe, M.: Weisfeiler and leman go neural: Higher-order graph neural networks. AAAI  (2019)

\bibitem{paszke2019pytorch}
Paszke, A., Gross, S., Massa, F., Lerer, A., Bradbury, J., Chanan, G., Killeen, T., Lin, Z., Gimelshein, N., Antiga, L., et~al.: Pytorch: An imperative style, high-performance deep learning library. NeurIPS  (2019)

\bibitem{peng2022fedni}
Peng, L., Wang, N., Dvornek, N., Zhu, X., Li, X.: Fedni: Federated graph learning with network inpainting for population-based disease prediction. IEEE TMI  (2022)

\bibitem{ramezani2021learn}
Ramezani, M., Cong, W., Mahdavi, M., Kandemir, M.T., Sivasubramaniam, A.: Learn locally, correct globally: A distributed algorithm for training graph neural networks. ICLR  (2022)

\bibitem{scardapane2020distributed}
Scardapane, S., Spinelli, I., Di~Lorenzo, P.: Distributed training of graph convolutional networks. IEEE Trans. Signal Inf. Process. over Networks  (2020)

\bibitem{sen2008collective}
Sen, P., Namata, G., Bilgic, M., Getoor, L., Galligher, B., Eliassi-Rad, T.: Collective classification in network data. AI magazine  (2008)

\bibitem{tripathy2020reducing}
Tripathy, A., Yelick, K., Bulu{\c{c}}, A.: Reducing communication in graph neural network training. SC  (2020)

\bibitem{wan2022bns}
Wan, C., Li, Y., Li, A., Kim, N.S., Lin, Y.: Bns-gcn: Efficient full-graph training of graph convolutional networks with partition-parallelism and random boundary node sampling. MLSys  (2022)

\bibitem{wang2020graphfl}
Wang, B., Li, A., Pang, M., Li, H., Chen, Y.: Graphfl: A federated learning framework for semi-supervised node classification on graphs. ICDM  (2022)

\bibitem{wu2021fedgnn}
Wu, C., Wu, F., Cao, Y., Huang, Y., Xie, X.: Fedgnn: Federated graph neural network for privacy-preserving recommendation. arXiv:2102.04925  (2021)

\bibitem{wu2022graph}
Wu, S., Sun, F., Zhang, W., Xie, X., Cui, B.: Graph neural networks in recommender systems: a survey. CSUR  (2022)

\bibitem{xie2021federated}
Xie, H., Ma, J., Xiong, L., Yang, C.: Federated graph classification over non-iid graphs. NeurIPS  (2021)

\bibitem{xu2018powerful}
Xu, K., Hu, W., Leskovec, J., Jegelka, S.: How powerful are graph neural networks? ICLR  (2019)

\bibitem{ying2018graph}
Ying, R., He, R., Chen, K., Eksombatchai, P., Hamilton, W.L., Leskovec, J.: Graph convolutional neural networks for web-scale recommender systems. KDD  (2018)

\bibitem{zeng2019graphsaint}
Zeng, H., Zhou, H., Srivastava, A., Kannan, R., Prasanna, V.: Graphsaint: Graph sampling based inductive learning method. ICLR  (2020)

\bibitem{zhang2021subgraph}
Zhang, K., Yang, C., Li, X., Sun, L., Yiu, S.M.: Subgraph federated learning with missing neighbor generation. NeurIPS  (2021)

\bibitem{zheng2021asfgnn}
Zheng, L., Zhou, J., Chen, C., Wu, B., Wang, L., Zhang, B.: Asfgnn: Automated separated-federated graph neural network. Peer Peer Netw. Appl.  (2021)

\end{thebibliography}

\appendix

\section{Algorithm}

We show the detailed steps in our proposed framework in \Cref{alg:model training}.

\begin{algorithm}[h]
    \caption{Our Proposed FGL Framework with Label-Guided Sampling}
    \label{alg:model training}
    \begin{tabular}{l}
    \makecell[l]{\textbf{Input}:
    normalized Laplacian matrices $\{\Tilde{\Mat{L}}_{ij}\}_{i,j=1,\dots,m}$; ini-\\tial weight matrices $\{\Mat{W}^l_0\}_{l=1,\dots,L}$; degree matrices $\{\Mat{D}_i\}$\\ for $i=1\dots m$; learning rate $\eta$; epoch number $T$; client\\ number $m$.} \\
         \textbf{Output}:
    Weight matrices $\{\Mat{W}^l_T\}_{l=1,\cdots,L}$.
    \end{tabular}
    \begin{algorithmic}[1]
        \FOR{$t=1$ to $T$}
        \FOR{$i=1$ to $m$ (in parallel)}
        \STATE Sample a local subset $\mathcal{V}_{S_i}$.
        \STATE \textbf{forward:}
        \FOR{$l=1$ to $L$}
        \STATE Calculate $\Mat{H}^l_{S_i}$; $\Mat{Z}^l_{S_i}=\sigma\left(\Mat{H}^l_{S_i}\right)$.
        \ENDFOR
        \STATE Calculate $Loss_{S_i}\left(\Mat{Z}_{S_i}^L,\Mat{W}\right)$.
        \STATE \textbf{backward:}
        \STATE Calculate $\nabla_{\Mat{Z}^L_{S_i}}\mathcal{L}_S$ locally.
        \STATE Calculate $\nabla_{\Mat{W}^L}\mathcal{L}_S$ on $V_S$.
        \FOR{$l=L$ to $1$}
        \STATE Calculate $\nabla_{\Mat{Z}^{l-1}_{S_i}}\mathcal{L}_S$ on $V_S$ and preserve.
        \STATE Calculate $\nabla_{\Mat{W}^{l-1}}\mathcal{L}_S$ on $V_S$.
        \ENDFOR
        \STATE $\Mat{W}^l_{t+1}\leftarrow\Mat{W}^l_t-\eta\nabla_{\Mat{W}^L}\mathcal{L}_S,\ l=1,\dots,L$
        \ENDFOR
        \ENDFOR
    \end{algorithmic}
\end{algorithm}

\section{Proof}
\textbf{Proposition 1.}
Matrix $\Mat{H}_{S_i}^l$ defined in \Cref{eq:subgraph estimator} is an unbiased estimator of hidden representation matrix $\Mat{H}_i^l$, i.e., $\E[\Mat{H}_{S_i}^l]=\Mat{H}_i^l$.
\begin{proof}
We first use zero rows to fill positions of nodes out of $\mathcal{V}_{S_i}$ to align the dimension of $\Mat{H}_{S_i}^l$ with $\Mat{H}_i^l\in\mathbb{R}^{n_i\times d^l}$.
According to the definition of $\Mat{H}_{S_i}^l$ in \Cref{eq:subgraph estimator}, consider independent sampling on different clients and we have
\begin{equation*}
\E[\Mat{H}^l_{S_i}\left[v,:\right]]=\sum_{j=1}^m\sum_{u\in\mathcal{V}_j}\frac{\Mat{L}_{ij}\left[v,u\right]}{p\left(u\right)}\Mat{Z}^{l-1}_j\left[u,:\right]\Mat{W}^l\E\left[\mathbb{I}_u\right],
\end{equation*}
where $\mathbb{I}_u=1$ if $u\in\mathcal{V_S}$, and $\mathbb{I}_u=0$, otherwise. 
Through the result $\E\left[\mathbb{I}_u\right]=p\left(u\right)$, it is evident that $\E[\Mat{H}_{S_i}^l]=\Mat{H}_i^l$.
\end{proof}

\noindent
\textbf{Theorem 1.}
Let $\E_T\|\nabla\mathcal{L}(\Mat{W}_T)\|_2=\frac{1}{T}\sum_{t=0}^{T-1}\|\nabla\mathcal{L}(\Mat{W}_t)\|_2$ where $T$ is the number of iterations. Assume $\mathcal{L}$ is $\rho$-smooth. Let $\eta\leq\frac{1}{\rho}$, we have $\E_T\Vert\nabla\mathcal{L}\left(\Mat{W}_T\right)\Vert_2^2\leq\frac{8\left(\mathcal{L}\left(\Mat{W}_0\right)-\mathcal{L}\left(\Mat{W}_T\right)\right)}{T\eta\left(2-\rho\eta\right)}$ with probability $1$.
\begin{proof}
Since the loss function $\mathcal{L}$ is $\rho$-smooth, the following inequality holds for the parameters $\Mat{W}_{t+1}$ and $\Mat{W}_{t}$
\begin{equation*}
\small
\mathcal{L}\left(\Mat{W}_{t+1}\right) \leq\mathcal{L}\left(\Mat{W}_t\right)+\nabla\mathcal{L}\left(\Mat{W}_t\right)^\top\left(\Mat{W}_{t+1}-\Mat{W}_t\right)+\frac{\rho}{2}\Vert\Mat{W}_{t+1}-\Mat{W}_t\Vert_2^2.
\end{equation*}
We then substitute $\Mat{W}_{t+1}=\Mat{W}_t-\eta\nabla{\mathcal{L}}_S\left(\Mat{W}_t\right)$ into the right-hand side. 
Given that $\E\left[\nabla{\mathcal{L}_S}\left(\Mat{W}_t\right)\right]=\nabla\mathcal{L}\left(\Mat{W}_t\right)+\left(\E\left[\nabla{\mathcal{L}_S}\left(\Mat{W}_t\right)\right]-\nabla\mathcal{L}\left(\Mat{W}_t\right)\right)$, we take the expectation value on both sides and arrive at
\begin{equation}
\small
\mathcal{L}\left(\Mat{W}_{t+1}\right)\leq\mathcal{L}\left(\Mat{W}_t\right)-(\eta-\frac{\rho\eta^2}{2})\Vert\nabla\mathcal{L}\left(\Mat{W}_t\right)\Vert_2^2+\left(\eta-\rho\eta^2\right)\delta\mathcal{L}_t\Vert\nabla\mathcal{L}\left(\Mat{W}_t\right)\Vert_2+\frac{\rho\eta^2}{2}\delta\mathcal{L}_t^2,
\end{equation}
where $\delta\mathcal{L}_t=\Vert\nabla\mathcal{L}\left(\Mat{W}_t\right)-\E\left[\nabla\mathcal{L}_S\left(\Mat{W}_t\right)\right]\Vert_2$ and $\eta\leq\frac{1}{\rho}$.
According to the strong law of large numbers, for any $v\in\mathcal{V}_{i}$, we have $\Mat{H}_{S_i}^l\left[v,:\right]\rightarrow\Mat{H}_i^l\left[v,:\right]$ 
as $s_i\rightarrow\infty$, resulting in $\delta\mathcal{L}_t\rightarrow 0$ with probability $1$, i.e., $\delta\mathcal{L}_t\leq\epsilon_0$ holds for a small enough $\epsilon_0>0$. 
In this sense, the above inequality turns to
\begin{equation*}
\small
\mathcal{L}\left(\Mat{W}_{t+1}\right)\leq\mathcal{L}\left(\Mat{W}_t\right)-(\eta-\frac{\rho\eta^2}{2})\Vert\nabla\mathcal{L}\left(\Mat{W}_t\right)\Vert_2^2+\epsilon_0\left(\eta-\rho\eta^2\right)\Vert\nabla\mathcal{L}\left(\Mat{W}_t\right)\Vert_2+\frac{1}{2}\rho\eta^2\epsilon_0^2.
\end{equation*}
By summing up all terms from $t=0$ to $T-1$ on both sides, we arrive at
\begin{equation*}
\small
\sum_{t=0}^{T-1}\Vert\nabla\mathcal{L}\left(\Mat{W}_{t}\right)\Vert_2^2\leq\frac{\epsilon_0\left(1-\rho\eta\right)}{1-\frac{1}{2}\rho\eta}\sum_{t=0}^{T-1}\Vert\nabla\mathcal{L}\left(\Mat{W}_{t}\right)\Vert_2+\frac{\mathcal{L}\left(\Mat{W}_0\right)-\mathcal{L}\left(\Mat{W}_T\right)}{\eta-\frac{1}{2}\rho\eta^2}+\frac{\rho\eta\epsilon_0^2T}{2-\rho\eta}.
\end{equation*}
According to the Cauchy-Schwartz inequality, we have $\sum_{t=0}^{T-1}\Vert\nabla\mathcal{L}\left(\Mat{W}_{t}\right)\Vert_2^2\geq T\cdot\E_T^2\|\nabla\mathcal{L}(\Mat{W}_T)\|_2$.
Consequently, we arrive at
\begin{equation*}
\small
\E_T^2\|\nabla\mathcal{L}(\Mat{W}_T)\|_2\leq\frac{\epsilon_0\left(1-\rho\eta\right)}{1-\frac{1}{2}\rho\eta}\E_T\|\nabla\mathcal{L}(\Mat{W}_T)\|_2+\frac{\mathcal{L}\left(\Mat{W}_0\right)-\mathcal{L}\left(\Mat{W}_T\right)}{T\eta(1-\frac{1}{2}\rho\eta)}+\frac{\rho\eta\epsilon_0^2}{2-\rho\eta}.
\end{equation*}
We next solve this quadratic inequality in terms of $\E_T\|\nabla\mathcal{L}(\Mat{W}_T)\|_2$ and get
\begin{equation*}
\small
\E_T\|\nabla\mathcal{L}(\Mat{W}_T)\|_2\leq\frac{\epsilon_0\left(1-\rho\eta\right)}{2-\rho\eta}+2\sqrt{\frac{\epsilon_0^2}{\left(2-\rho\eta\right)^2}+\frac{\mathcal{L}\left(\Mat{W}_0\right)-\mathcal{L}\left(\Mat{W}_T\right)}{T\eta\left(1-\frac{1}{2}\rho\eta\right)}},
\end{equation*}
for $\eta\leq\frac{1}{\rho}$. 
When $\varepsilon_0$ is small enough, we can approximately neglect the $\epsilon_0$ terms and finish the proof.
\end{proof}

\section{Complexity Analysis}
Our proposed label-guided sampling approach balances the distribution in the sampled subgraph to mitigate the local bias. At the same time, it significantly reduces training overhead. To better demonstrate its effectiveness in scalability, we analyze training overhead, e.g., communication, computation, and memory overhead in this subsection.

In the forward stage, each client $C_j$ sends $m-1$ local terms $\Tilde{\Mat{L}}_{S_{ij}}\Mat{O}_{S_j}^{l-1}$ for $i\neq j$ to the central server. Then, for client $C_i$, the central server sums up the local terms \wrt. $C_i$ from all other clients as $SUM_i=\sum_{j\neq i}\Tilde{\Mat{L}}_{S_{ij}}\Mat{O}_{S_j}^{l-1}$. Finally, the central server sends $SUM_i$ to the client $C_i$, and consequently $\Mat{Z}_{S_i}^l$ is calculated within $C_i$. The total communication volume for model inference is $mn_SLd$ per iteration. 

In the backward stage, the computation of $\nabla_{\Mat{Z}^L_{S_i}}{\mathcal{L}_S}$ does not require any communication because $\nabla_{\Mat{Z}^L_{S_j}}Loss_{S_i}=0,\ i\neq j$. For the computation of $\nabla_{\Mat{Z}^{l-1}_i}\mathcal{L}_S$, we follow the same communication steps as inference with $\Tilde{\Mat{L}}_{S_{ij}}\Mat{P}_{S_j}^{l}$ as local terms.
For the computation of $\nabla_{\Mat{W}^l}\mathcal{L}_S$, we save the result $\nabla_{\Mat{Z}^{l-1}_{S_i}}\mathcal{L}_S$ locally. Each client $C_i$ only sends the local gradient $(\Mat{Z}^{l-1}_{S_i})^\top\nabla_{\Mat{Z}^{l-1}_{S_i}}\mathcal{L}_S$ to the central server. Then the central server sums them up and scatters the global summation to all clients to update the local models. 
The computation of $\nabla_{\Mat{Z}^{l-1}_i}\mathcal{L}_S$ and $\nabla_{\Mat{W}^l}\mathcal{L}_S$ requires $mLd(n_S+d)$ communication volume in total.
Assuming that the element-wise activation $\sigma(\cdot)$ can be ignored, the computational time required is $2Ln_Sd^2+n_Sd$.

In addition, the memory overhead includes two parts, the feature matrix, and the model parameters. The size of the feature matrix is $n_Sd$, and $n_SLd$ in total for $L$ layers. The model parameters are kept in each client and require $mLd^2$ of memory in total. Therefore, the overall memory overhead is $n_SLd+mLd^2$.

Moreover, we observe that BNS-GCN~\cite{wan2022bns} has a similar overhead as ours. 
In particular, BNS-GCN reduces communication overhead by introducing redundant memory of boundary node features in different devices. In fact, BNS-GCN can be seen as a different sampling distribution from ours where $p_v=1$ when $v\in\sB$ and $p_v=\frac{1}{n_S}$ otherwise. 

\section{Experimental Settings}
\begin{table*}[t]
\centering
\tabcolsep = 2 pt
\caption{Statistics and details of three datasets for node classification.}\label{tab:dataset}
\begin{tabular}{lccccc}
    \toprule[1pt]
    Dataset & Nodes & Edges & Features & Classes & Train / Val / Test \\
    \midrule[0.5pt]
    PubMed & 19,717 & 44,338 & 500 & 3 & 18,217 / 500 / 1,000 \\
    Reddit & 232,965 & 11,606,919 & 602 & 41 & 153,431 / 23,831 / 55,703 \\
    Products & 2,449,029 & 123,718,280 & 100 & 47 & 196,615 / 39,323 / 2,213,091 \\
    \bottomrule[1pt]
\end{tabular}
\end{table*}

\subsection{Datasets}
We adopt three prevalent datasets PubMed \footnote{https://linqs.soe.ucsc.edu/data}, Reddit \footnote{http://snap.stanford.edu/graphsage/} and Ogbn-products \footnote{https://github.com/snap-stanford/ogb} in previous works~\cite{sen2008collective,hamilton2017inductive,hu2020open} to evaluate our proposed framework. PubMed is a citation network with nodes as scientific papers and edges as citations; Reddit is a large social network consisting of Reddit posts in different communities, with nodes as posts, and two posts share an edge when they are commented on by the same Reddit user; Ogbn-products is an Amazon product co-purchasing network where nodes represent products and edges indicate that the products are purchased together. We present details of these datasets in \Cref{tab:dataset}.

\subsection{Baselines}
We compare our proposed framework with existing methods in centralized and distributed settings. For centralized methods, we have
\begin{itemize}[leftmargin=*]
    \item GCN \cite{kipf2016semi} is a specific type of GNN that introduces the spectral graph convolution operation to GNNs.
    \item 1-GNN \cite{morris2019weisfeiler} is a generalization of GCN that enhances self-connection with a linear map.
    \item GraphSAGE \cite{hamilton2017representation} is a specific type of GNN that develops different neighbor aggregation operations and a node sampling method for mini-batch training.
    \item GraphSAINT \cite{zeng2019graphsaint} is an inductive GCN with subgraph sampling and normalization.
\end{itemize}
For distributed methods, we have
\begin{itemize}[leftmargin=*]
    \item FedAvg+GCN is an FGL framework that combines GCN \cite{kipf2016semi} and FedAvg \cite{mcmahan2017communication} to learn a shared global model.
    \item CAGNET \cite{tripathy2020reducing} is a distributed GCN training framework where computation is conducted on small client groups in parallel.
    \item BNS-GCN \cite{wan2022bns} is a distributed GCN framework with a novel boundary node sampling technique.
    \item FedSAGE+ \cite{zhang2021subgraph} is an FGL framework that exploits a missing neighbor generator to approximate the cross-client neighbors in the local training.
    \item LLCG \cite{ramezani2021learn} is a distributed GNN training framework that leverages a global server correction method to reduce local bias in the local training step.
\end{itemize}

\subsection{Implementation}
We implement our framework and existing baselines with Pytorch \cite{paszke2019pytorch} on one Intel(R) Xeon(R) CPU E5-2660 v4 @ 2.00GHz with 28 CPU cores. To simulate a distributed learning environment, we use the communication primitives of the "Gloo" backend in the \emph{torch.distributed} package. We use Adam \cite{kingma2014adam} as an optimizer for training.
Our code is available at \url{https://anonymous.4open.science/r/LocalBias-26B1}.

To reproduce our experimental results, we fix the dimension of hidden features as 128, and the probability of dropout as 0.2. We independently search for the optimal learning rate among 0.1, 0.01, and 0.001 in each case. For all methods, the number of the backbone layer is 2 in PubMed, 4 in Reddit, and 3 in Ogbn-products; the number of epochs is 200 in PubMed and Reddit, and 1,000 in Ogbn-products. Specifically, for distributed baselines, we set the number of clients as 8 to ensure a fair comparison. We start 8 processes to train our FGL framework in parallel. We randomly partition all nodes into 8 subsets with the same number of nodes and distribute each subset with its feature matrix, label vector, and edge set to one process. Note that we use random partition instead of classical graph partition algorithms, e.g., Metis~\cite{doi:10.1137/S1064827595287997} because the data itself is produced in a distributed manner and we cannot arbitrarily control the proportion of cross-client edges. We fix the size of sampled subgraph as 5,000 for PubMed, 50,000 for Reddit, and 500,000 for Ogbn-products. 
All experiments are conducted in a transductive learning setting.
In addition, we list some key packages in Python required for implementing certified unlearning as follows.
\begin{itemize}[leftmargin=*]
\item python == 3.9.17
\item torch == 2.0.1
\item torch-geometric == 2.3.1
\item ogb == 1.3.6
\item numpy == 1.21.5
\item scikit-learn == 1.3.0
\end{itemize}

\section{Supplementary Experiments}

\subsection{Parameter Study}

\begin{wrapfigure}[12]{r}{0.4\textwidth} 
\centering
\vspace{-2mm}
\includegraphics[width=\linewidth]{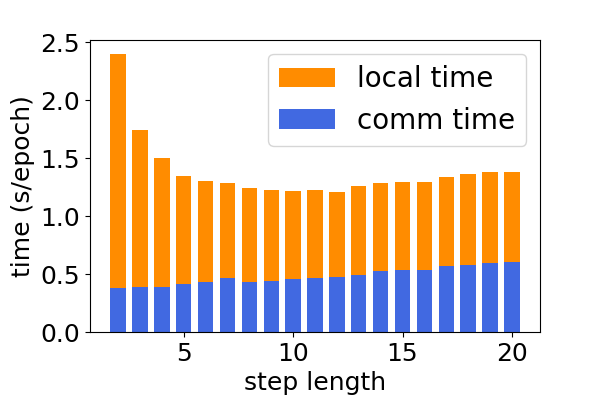}
\vspace{-6mm}
\caption{Local computation time and communication overhead with different sample sizes on Reddit.}
\label{fig:subgraph_size}
\end{wrapfigure}

In the experiments, we observe that the size of sampled subgraph in our global GNN training has a significant effect on model efficiency. We further conduct experiments with different sample sizes and present the results in \Cref{fig:subgraph_size}. 
For each epoch, we fix the number of nodes in the training set as 100,000; when the step length is 2 for instance, the batch size is computed as 100,000 / 2 = 50,000 correspondingly, and model parameters are updated twice per epoch. After training for 10 epochs, we obtain the average communication cost and local computation time. \Cref{fig:subgraph_size} illustrates that as we set a larger step length, communication time increases steadily and local computation time decreases with a decaying rate. In particular, the total training time reaches a minimum when the step length is set to 12. This indicates that the step length (and consequently batch size) governs the trade-off between local computation and communication overhead, where the optimal value varies in different experiment settings.

\subsection{Convergence Study}
\begin{wrapfigure}[10]{r}{0.6\textwidth} 
\centering
\vspace{-11mm}
\subfigure[Reddit]{
\includegraphics[width=0.47\linewidth]{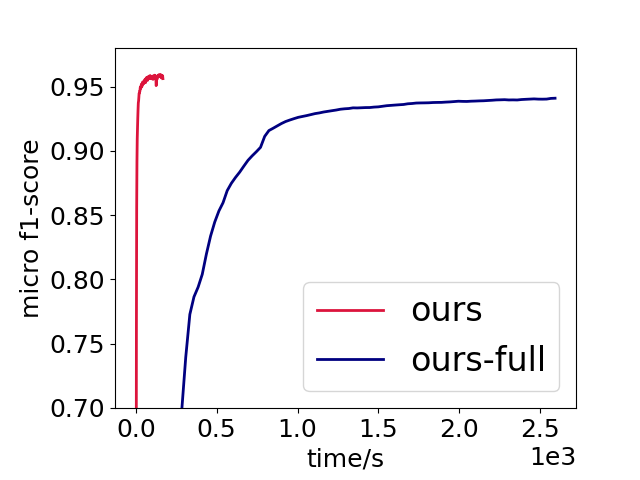}}
\subfigure[Ogbn-Products]{
\includegraphics[width=0.47\linewidth]{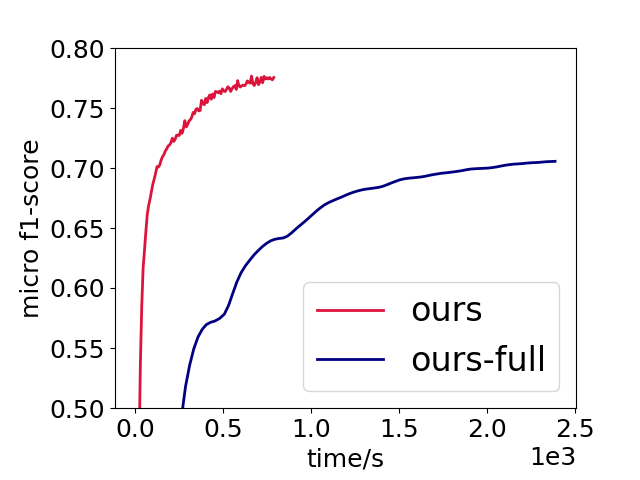}}
\vspace{-4mm}
\caption{Test set convergence curves of ours and ours-full on two datasets.}
\label{fig:accuracy_time}
\end{wrapfigure}
We find that our proposed training algorithm also has a fast convergence rate in addition to a low training overhead. 
Both points are significant to the overall efficiency of training. 
To verify the fast convergence of the training algorithm, we record the micro-F1 score and the training time in each epoch. 
In \Cref{fig:accuracy_time}, we illustrate the convergence curves of our method with label-guided sampling (ours) and the full batch variant (ours-full), namely, the change in the micro-F1 score with training time. 
It is demonstrated that our framework requires significantly less time to converge while achieving similar or even better performance. 
This indicates that our label-guided subgraph sampling method greatly contributes to rapid convergence.

\subsection{Ablation Study}
\begin{table}[t]
\centering
\tabcolsep = 6 pt
\caption{Ablation study of graph sampling and normalization in our framework where ours-full stands for dropping the sampling and ours* denotes dropping the normalization.}
\label{tab:norm}
\begin{tabular}{lccccc}
    \toprule[1pt]
    Method & Sampling & Normalization & PubMed & Reddit & Products \\
    \midrule[0.5pt]
    ours-full & $\times$ & $\times$ & 87.60\% & 94.63\% & 74.10\% \\
    ours* & $\checkmark$ & $\times$ & 89.70\% & 95.89\% & 78.04\% \\
    ours & $\checkmark$ & $\checkmark$ & \textbf{90.10}\% & \textbf{96.05}\% & \textbf{79.15}\% \\
    \bottomrule[1pt]
  \end{tabular}
\end{table}
In our experiments, we observe that sampling and normalization greatly impact the performance of our framework. 
Thus, we design ablation experiments to remove them separately and run experiments with ablated versions. The results in \Cref{tab:norm} demonstrate that our framework (sampling and norm) performs best in most cases. And ours* (only sampling) has a bit lower performance, but is still better than ours-full (no sampling). Note that the normalization coefficient satisfies $\frac{1}{\alpha}>1$, hence ours has a larger variance in performance than ours*. In addition, we find that normalization largely accelerates the convergence of our training algorithm. 
As a result, both sampling and normalization steps are crucial to our method's performance.

\end{document}